\documentclass[runningheads]{llncs}

\usepackage{eccv}

\usepackage{eccvabbrv}

\usepackage{graphicx}
\usepackage{booktabs}

\usepackage[accsupp]{axessibility}  

\usepackage{hyperref}

\usepackage{orcidlink}

\begin{document}

\title{Non-verbal Interaction and Interface with a Quadruped Robot using Body and Hand Gestures: Design and User Experience Evaluation}

\titlerunning{Non-verbal Interaction with a Quadruped Robot}


\author{Soohyun Shin\inst{1}, Trevor Evetts\inst{2} \and Hunter Saylor\inst{3} \and Hyunji Kim\inst{4} \and \\ Soojin Woo \inst{1} \and Wonhwha Rhee \inst{1} \and  Seong-Woo Kim\inst{1} 
\thanks{The authors 2, 3, 4 participated in this research during their visiting research at Autonomous Robot Intelligence Lab (ARIL), SNU, https://aril.snu.ac.kr}.}

\authorrunning{S. Shin et al.}

\institute{Seoul National University, Seoul, Korea \\
\email{\{soohyunshin,soojin.woo,legend123,snwoo\}@snu.ac.kr} \and
University of Florida, Gainesville, Florida, USA \\
\email{trevor.evetts@ufl.edu} \and
Morgan State University, Baltimore, Maryland, USA \\
\email{husay1@morgan.edu} \and
University of Toronto, Toronto, Ontario, Canada \\
\email{hji.kim@mail.utoronto.ca}}

\maketitle


\begin{abstract}
In recent years, quadruped robots have attracted significant attention due to their practical advantages in maneuverability, particularly when navigating rough terrain and climbing stairs. As these robots become more integrated into various industries, including construction and healthcare, researchers have increasingly focused on developing intuitive interaction methods such as speech and gestures that do not require separate devices such as keyboards or joysticks. This paper aims at investigating a comfortable and efficient interaction method with quadruped robots that possess a familiar form factor. To this end, we conducted two preliminary studies to observe how individuals naturally interact with a quadruped robot in natural and controlled settings, followed by a prototype experiment to examine human preferences for body-based and hand-based gesture controls using a Unitree Go1 Pro quadruped robot. We assessed the user experience of 13 participants using the User Experience Questionnaire and measured the time taken to complete specific tasks. The findings of our preliminary results indicate that humans have a natural preference for communicating with robots through hand and body gestures rather than speech. In addition, participants reported higher satisfaction and completed tasks more quickly when using body gestures to interact with the robot. This contradicts the fact that most gesture-based control technologies for quadruped robots are hand-based. The video is available at \url{https://youtu.be/rysv1p1zvp4}.
\end{abstract}


\section{Introduction}
\label{sec:intro}

In recent years, the quadruped robot has received significant attention due to its many pragmatic and maneuverable benefits. Resembling a four-legged animal, they are able to maneuver through rough terrain and climb stairs with more stability compared to robots with other means of mobility. In parallel with these hardware developments, studies in the multimodal human-computer interface (MHCI) have grown considerably, with researchers exploring how to collaborate on various tasks with such robots in an intelligent way\cite{halder2023construction}. Although there are questions to be answered on how people will engage with robots in an unprompted context, such as on the street \cite{chan2023understanding}, there are just as many questions regarding the best way for people to control robots in a prompted context, with time to learn and adapt to the controls in front of them \cite{chappuis2024learning}.

Gesture-based control requires a synthesis of both technological development and human understanding \cite{song2012continuous}. On the one end, natural user interfaces seek to give the user a pleasurable experience in controlling the robot, whether through voice, with a controller, etc. \cite{li2020human}. Meanwhile, hardware and software must be capable of understanding and interpreting the inputs given by a human. Therefore, engineers and scientists must understand how humans want to engage with certain methods of control while adapting to the technical challenges these human needs might require.

The goal of this research is to understand the factors in gesture user interface (UI) design to control quadruped robots and to evaluate how users perceive controlling the robot through the gesture UI. This study investigates the effectiveness of body versus hand gestures in controlling a quadruped robot. Through a series of experiments, we assessed user satisfaction and task efficiency using these different interaction methods. We first conducted two preliminary studies that observed how people interact with the quadruped robot in both a natural and controlled environment using the Go1 Pro quadruped robot, in which we learned how people intuitively tried to communicate with the robot using hand and body gestures. Then we designed and built a prototype system that recognizes user movements that can be controlled using the Google Mediapipe for pose detection and classification. Finally, we conducted an experiment in which 13 participants directed the quadruped robot go through a zigzag shaped line using the gesture we designed and evaluated their user experience with User Experience Questionnaire (UEQ) \cite{schrepp2017construction} and measured the time taken to complete the tasks.

The preliminary results suggest a natural human inclination toward interacting with robots through hand and body gestures, rather than speech. Moreover, participants demonstrated slightly higher satisfaction and completed tasks slightly more quickly when employing body gestures to control the robot. This observation challenges the predominant focus of most gesture-based control technologies for quadruped robots, which are primarily hand-based. Additionally, we will discuss the insights and considerations derived from our findings.

To sum up, the contributions of this paper is as follows:
\begin{itemize}
\item Descriptive findings from two preliminary studies on how people interact with quadruped robots in a natural and controlled setting.
\item A prototype of a system that recognizes human hand and body movement to control the quadruped robot.
\item Quantitative and qualitative results of a user experience study of 13 participants controlling the quadruped robot with hand and body gestures.
\end{itemize}

\section{Related Works}

\subsection{Human Interaction with Quadruped Robots}
Recent studies have tried to understand how humans interact with the quadruped robot in both natural and controlled settings. Since the emergence of InCoPs (incidentally copresent persons) research \cite{rosenthal2020forgotten}, there have been many attempts to understand the social and psychological perception of robots from the perspective of daily pedestrians \cite{hardeman2021encounters}. Previous research has shown that how the robot looks and acts greatly affects the user's perception of the robot \cite{goetz2003matching, liu2022friendly}. Therefore, the interaction design of the quadruped robot with pedestrians should take into account the size, appearance and speed of the robot as well as its personality design \cite{hashimoto2024safe}. The Go1 Pro quadruped that we used for our experiments resembles a small dog and seemed to be unthreatening and adorable, as we will reveal in the later parts of this paper. There have also been attempts to understand how to collaborate with quadruped robots in environments for specific industries such as construction \cite{halder2023construction}, healthcare \cite{cai2024navigating}, and agriculture \cite{ferreira2022survey}.

\subsection{Gestural User Interface Design}
Several studies have designed and tested gesture-based user interfaces for controlling robots. In particular, there have been many studies on the implementation and testing of the user experience of hand and body gesture control of drones \cite{yam2022commanding, gio2021control, bhat2021real}, because drones were widely adopted before quadruped robots. As such, previous studies have shown that using natural interfaces such as gestures can be useful to control a robot. Unlike interacting with a computer screen, interacting with a robot can involve multimodal interaction such as speech or gesture. While significant advancements have been made in gesture-based control systems for robots, few studies have specifically focused on the effectiveness of body gestures in controlling quadruped robots \cite{qi2024computer}. This research aims to fill this gap by comparing the user experience of hand and body gesture. Additionally, previous research shows a design methodology for robot interaction, such as bodystorming \cite{porfirio2019bodystorming}. This is a method in which researchers act out an encounter with the robot to explore how they would act if they meet a robot. Insprired by this method, we conducted preliminary studies that observe people's gestures as they encounter the quadruped robot. 

\subsection{Gesture-Based Control Systems}
There have been various studies exploring the use of gestures in Human-Robot Interaction (HRI) to control quadruped robots. These studies include efforts to improve hand gesture recognition accuracy \cite{zafar2024real}, enable distance recognition \cite{bamani2024ultra}, use multimodal sensors, including surface electromyography \cite{uimonen2023gesture}, and detect hand gestures from remote users without relying on robot sensors \cite{li2023gesture}. All of these studies focus primarily on hand shapes and aim to enhance recognition accuracy.

Our research takes a different approach. The fundamental question in our study is how people perceive interaction with robots that have a somewhat familiar animal-like form and which methods of interaction are most comfortable and effective in accomplishing assigned tasks and missions. The aim of this paper is to explore how robots that interact with and co-exist with humans can communicate effectively with people, which we believe will ultimately contribute to enabling collaboration and teamwork between humans and robots.

\section{Experiment}
\subsection{Preliminary Studies}

\subsubsection{Experimental setup}
Before the main experiment, we conducted two preliminary studies to first see how people naturally interact with the quadruped robot. The first study was conducted at various points of the campus, and the other study was conducted in the hallway outside of our lab. We used a Go1 Pro quadruped robot for the experiment. Out of the box, it does not move autonomously other than that it can follow the person who has the tracker. The robot can be controlled with a controller and runs on a battery that can run for about 2 hours. The shape of the robot resembles a dog and its size is relatively small, and its dimensions are 0.588 by 0.22 by 0.29 meters \cite{unitree}. With the controller, the Go1 Pro can walk, run, and jump, stand up and even dance. 


\begin{figure}[tb]
  \centering
  \includegraphics[height=3.1cm]{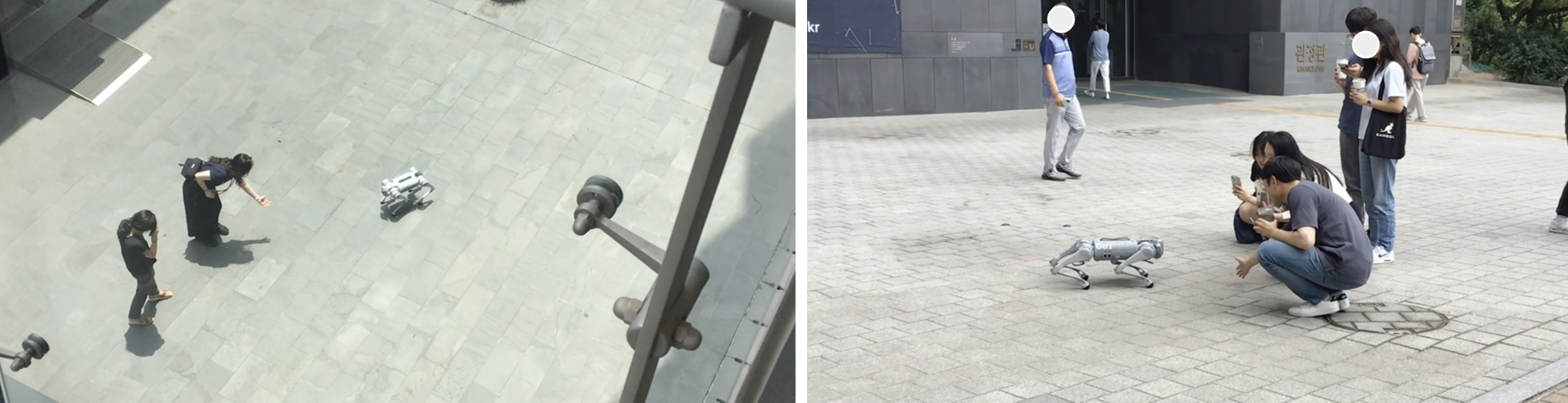}
  \caption{Pedestrians on the campus attempting to engage with the Go1 Pro.
  }
  \label{fig:study1}
\end{figure}

The preliminary studies were conducted in the "Wizard of Oz" style (WoZ), without any programming adjustments to the robot \cite{rietz2021woz4u, riek2012wizard}. This type of experiment is a research methodology commonly used in the fields of human-computer interaction (HCI) and human-robot interaction (HRI) to test and evaluate user interactions with a system that is not yet fully functional or automated. In such an experiment, participants interact with a system or robot that they believe to be autonomous, but which is actually being controlled or manipulated by a human operator hidden from the participants' view. This setup allows researchers to simulate and study complex interactions and gather valuable user feedback without needing a fully operational system. In our study, instead of the robot interpreting the participants' gestures autonomously, one of our researchers acted as a hidden human operator that observed the gestures through a camera and manually controlled the robot to respond accordingly. This setup allows us to study how intuitive the gesture controls are, how users expect the robot to behave, and what challenges might arise in real-world interactions, all without needing the robot to have fully functional gesture recognition capabilities. Below are the details of how each study was conducted and our findings. 

\subsubsection{Study 1: Observation of Interaction with Pedestrians}
Our first study was conducted by taking the quadruped robot to parts of the university campus where this research was carried out, with the objective of learning people's natural gestural reaction when encountering the robot for the first time. We chose several locations for the study, including a street that many people passed by and a three-way intersection that many people gathered around. One of our researchers controlled the robot with a controller device from a hidden place, which pedestrians cannot easily see. Then, when a few pedestrian tried to interact with the robot, our researcher made the robot dance or stand up so that it would engage with people. The nature of this experiment was more playful, as we engaged with the passerby until they chose to move away. Figure \ref{fig:study1} shows the interaction between the quadruped and the pedestrians. 


As a result of this first study, we found that most people's attitudes were friendly to the robot. Many people took photos with their smartphone. While some people seemed scared and walked away, most people immediately treated the robot as an actual dog, offering their hand and tried to pet its head. We found that just as interacting with a real dog, people stooped down to the robots' "eye" level to engage with it. Our findings showed that people used many body gestures to engage with the robot. However, since pedestrians did not have a specific reason to interact with the robot, we found that most interactions were shallow and without purpose. Therefore, we decided to conduct a second study in a more controlled environment and give each participant a specific task to accomplish with the GO 1 Pro. 

\begin{figure}[tb]
  \centering
  \includegraphics[height=6cm]{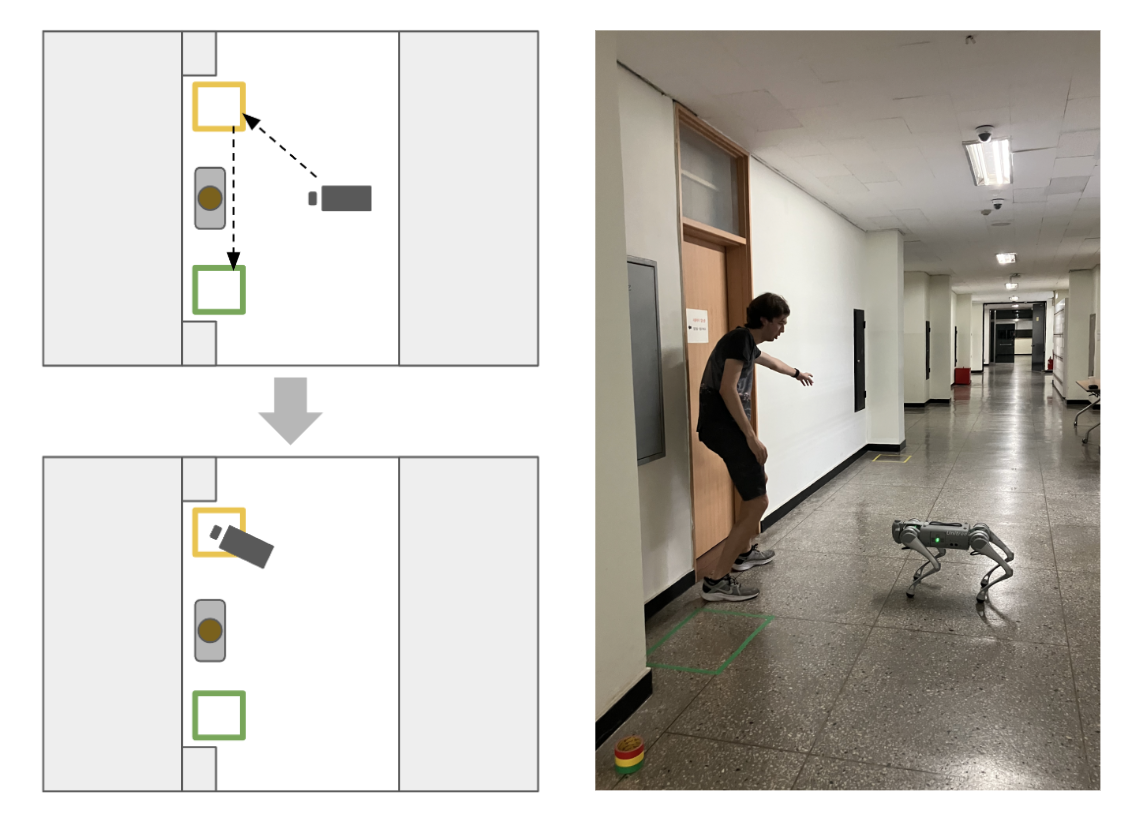}
  \caption{Diagram of first part (task 1) of in lab study, and a scene from the experiment.   
  }
  \label{fig:study2_1}
\end{figure}

\subsubsection{Study 2: Observation of Interaction with Users Given a Task}
Our second study took place in the hallway outside of our laboratory. Five people in our lab participated in this research. As mentioned above, our purpose in this study was to observe how users interact more purposefully with the quadruped robot in a controlled environment with a specific task. In a hallway about 10 meters long, we marked two spot with different colors between two beams in the hallway. The users were assigned the task of maneuvering the robot in the yellow spot using only their hand and body gestures as input. Figure \ref{fig:study2_1} shows the experimental setting. Like in the first study, one of our researchers controlled the robot from a hidden place so that the participants felt as if they were interacting with an autonomous robot. 


\begin{figure}[tb]
  \centering
  \includegraphics[height=3cm]{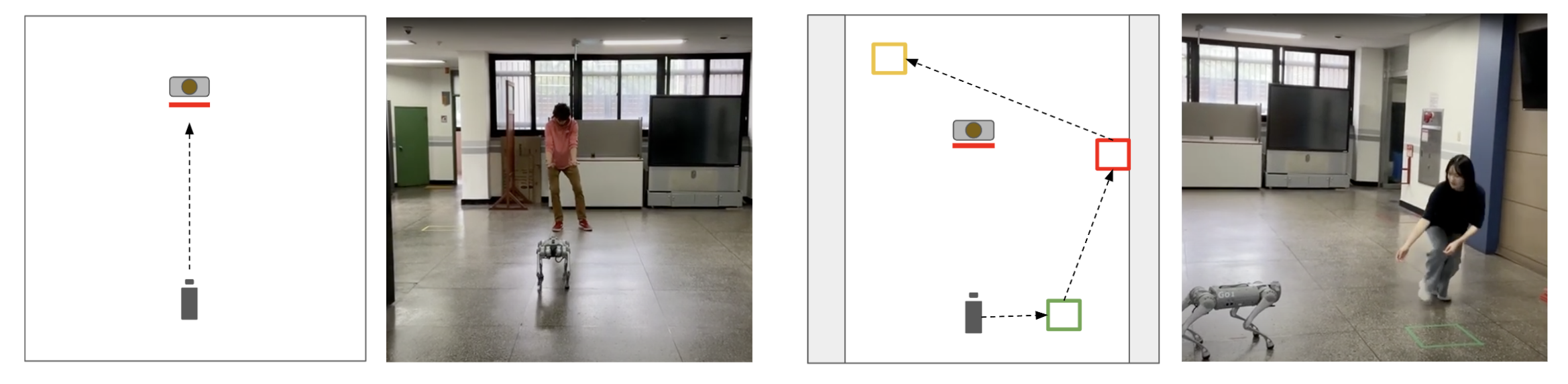}
  \caption{Diagram of second part (task 2 and task 3) of in lab study and a scene from the experiment.}
  \label{fig:study2_2}
\end{figure}

We wanted to find out if there are additional robot actions we should consider when designing human interaction with the quadruped, so we devised a more complex environment where we marked three spots each with a different color. The spots were placed in a triangle. Then we instructed each participant with two tasks that they needed to complete. Figure \ref{fig:study2_2} shows the experimental setting. The following is a summary of each experiment and our findings. 
\\ \\
\textbf{Task 1}: Command the Go1 Pro to move to the yellow spot, then to the green spot.
\\ \\
\textbf{Task 2}: Get Go1 Pro’s attention, and send gesture signal that you want to interact. Move the Go1 Pro to the red line. Participant cannot move from the spot.
\\ \\
\textbf{Task 3}: Get the Go1 Pro’s attention, gesture the Go1 Pro to move from green, to red, to yellow. Participant should start from the red line, but s/he can move to wherever once the study starts.   
\\ \\
\textbf{Lessons learned}: 
\begin{itemize}
\item As soon as the Go1 Pro runs into a corner, it can no longer see the gestures. Unlike drones, the quadruped's head turns away from the user, thus hindering the robot to view and receive any further directions from the user. One way to solve this is make the robot turn to face the user once the command is completed. 
\item The participant's natural instinct is to move to the spot and point downward. 
\item The most effective way to engage is to interact with someone positioned in front of the robot. 
\end{itemize}
From the observations, we were able to design the user experience for our final experiment, where we evaluated the user experience of each participant with UEQ. 


\section{Prototype Experiment and User Experience Evaluation}

\subsection{System Implementation}
\begin{figure}[t]
  \centering
  \includegraphics[width=\textwidth]{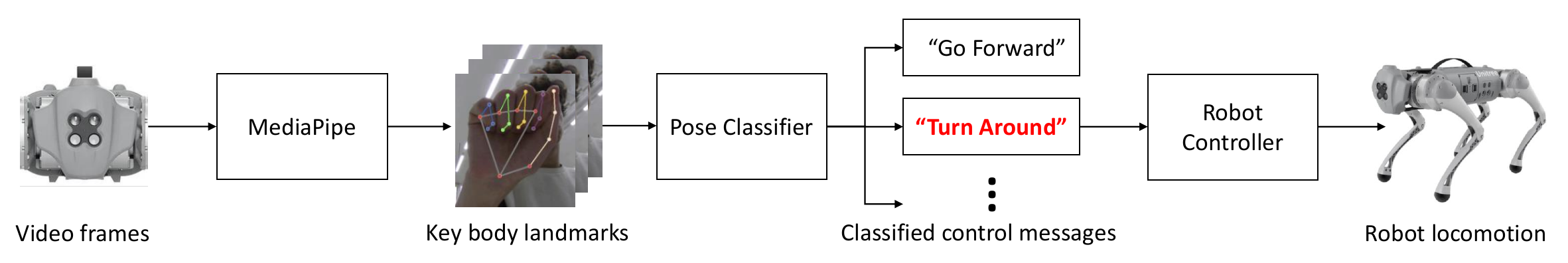}
  \caption{Software architecture of gesture-based quadruped robot control system.
  }
  \label{fig:software_architecture}
\end{figure}


Based on the results from the previous studies, we designed and developed a quadruped robot capable of directly recognizing and responding to human gestures. This autonomous system operates independently, using data from its own sensors without external intervention. All subsequent experiments were conducted using this robot. 

The primary consideration in implementing the gesture-interactive quadruped robot was pose estimation and classification. Our prototype required models and methods that balanced intuitiveness, affordability, and performance speed. We explored the OpenPose model, which can track multiple users simultaneously but has specific CPU and GPU requirements, highlighting the need for less resource-intensive alternatives.

Mediapipe, developed by Google, offers a fast and budget-friendly solution for pose detection. It supports multimodal input and can identify 33 key body landmarks, including joints and points of interest such as the eyes and hands. Although Mediapipe has limitations, it is suitable for pose-related experiments. Pose classification can be achieved through trigonometric calculations of limb angles, reducing computational demands and allowing quick Python-based calculations, without ROS (Robot Operating System) or extensive ML processes. 

Figure \ref{fig:software_architecture} shows the software architecture of our system. Mediapipe can be used to identify and model key points in the body and hands of users. It can even estimate the user’s position in 3D space, allowing for graphical models of a person’s figure. The ability to map these coordinates on a Cartesian plane is the backbone of the pose classification software that our team used. 

\begin{figure}[t]
  \centering
  \includegraphics[width=\textwidth]{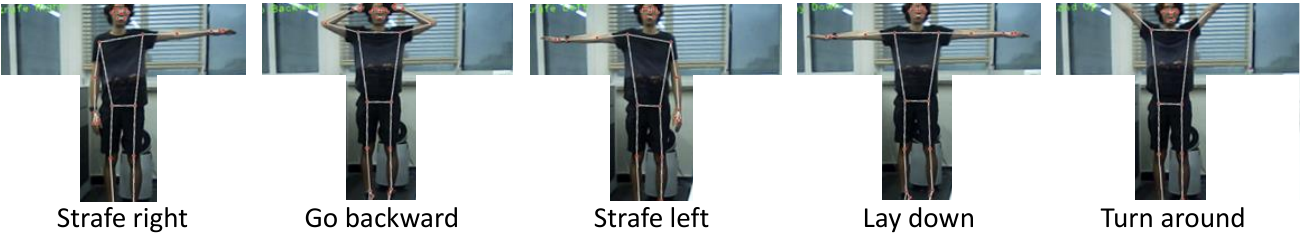}
  \caption{The recognized pose of body gestures. Note that the images have been mirrored and rotated four degrees counterclockwise for better visibility for the readers.}
  \label{fig:body_gesture}
\end{figure}

When a user creates a certain pose, angles form in the space between their limbs. Let the line between the shoulder and elbow of a person be vector AB. Furthermore, let the line between someone's shoulder and hip be vector AC. Therefore, the interior angle to these two vectors will be poses, by nature, result in (or are the product of) unique configurations of these angles. If a user could account for each angle made by joints, they could classify any variation of unique poses. We used a total of eight angles associated with the human body; two shoulders, two elbows, two hips, and two knees. With angles ranging from zero to 360 degrees, the software can classify a pose if the angles fall into the appropriate ranges. Although Mediapipe is a machine learning algorithm for identifying key points, pose classification does not require modeling training. Bounds can be adjusted to make it easier or more difficult for the user to achieve certain poses; make a bound too large, and a false positive can occur. However, make a bound too small, and a user could struggle to successfully make the pose. Figure \ref{fig:body_gesture} provides examples of the recognized pose of body gestures.

\begin{figure}[t]
  \centering
  \includegraphics[width=\textwidth]{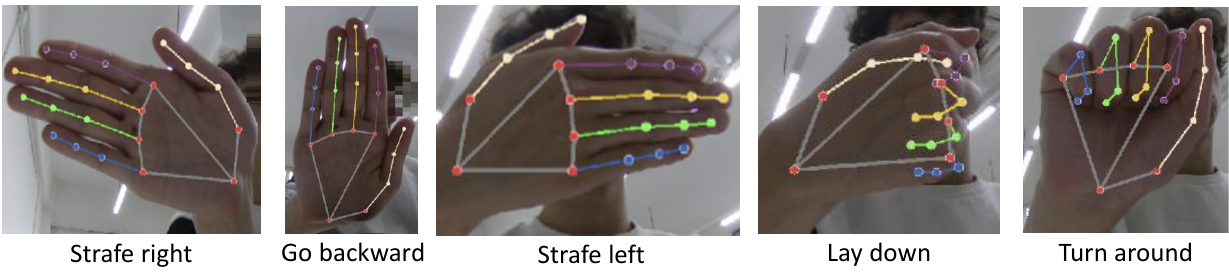}
  \caption{The recognized key landmarks of hand gestures. Note that images are mirrored.}
  \label{fig:hand_gesture}
\end{figure}

The hand pose classification works differently from the body pose classification. Hand pose-classification takes the keypoints from the hand, and then maps each keypoint on an xy-plane. Then, to classify the poses, Mediapipe compares the x coordinates and y coordinates of the relevant key points to determine the position and orientation in which the user's hand is in. For example, to see if the user's index finger is pointed upward, Mediapipe compares the y coordinate of the tip of the index finger with the y coordinate of the knuckle of the index finger, and if the y coordinate of the fingertip is greater than that of the knuckle, then the user is likely to point their index finger upward. Figure \ref{fig:hand_gesture} provides examples of the recognized hand key landmarks of hand gestures.

\subsection{Gesture Design and Recognition}
Given the findings from previous studies, which predominantly focus on hand gestures, our study explores both hand and body gestures through a series of controlled experiments designed to evaluate user experience comprehensively. We devised hand and body gestures that would be easily recognized by our system. Figure \ref{fig:hand_gesture} presents the list of nine hand and body gestures we designed. Figures \ref{fig:body_gesture} and \ref{fig:hand_gesture} provide examples of the recognized key landmarks for 5 of the 9 gestures in Figure \ref{fig:hand_gesture}. For clear communication, our system waits two seconds after the robot stops moving before issuing the next command. Also, note that the camera images are mirrored. 


\begin{table}[t]
\caption{Hand and body gesture design.}
  \label{fig:hand_body_gesture_table}
\centering
\begin{tabular}{|l|l|l|} \hline
\textbf{Action} & \textbf{Hand Gesture} & \textbf{Body Gesture} \\ 
\hline \hline
Go forward & Point up & Hands on hips  \\ 
\hline
Go backward & Palm out/High five & Hands on head \\ 
\hline
Rotate counter-clockwise & Point left & Left arm bent \\ 
\hline
Rotate clockwise & Point right & Right arm bent\\ 
\hline
Strafe left & Sideways left & Left arm out\\ 
\hline
Strafe right & Sideways right & Right arms out \\ 
\hline
Lay down & Fist left & T-pose \\ 
\hline
Stand Up & Fist right & Arms elevated \\ 
\hline
Turn Around & Fist & Both arms bent \\ 
\hline
\end{tabular}
\end{table}

\subsection{Method}
\subsubsection{Participants}

The experimental portion of our study included 13 participants, ranging in age from 20 to 40 years, from South Korea and Canada, and the United States. While some of the participants had experience with robotics, none had engaged with the interfaces that we had designed for them to interact with before the experiment. Using the experiment, we sought to test the feasibility of our pose and hand recognition technology with these study participants, as well as to compare the two ways of controlling the robot. The sample size of this study aligns with similar exploratory studies in the field of Human-Robot Interaction (HRI), where small group participants are often utilized to gather preliminary information and identify key trends before conducting larger-scale research. Due to the resource-intensive nature of setting up and conducting experiments with robotics, particularly those involving gesture recognition and control systems, this study focused on a more manageable sample size that allowed for detailed observation and interaction with each participant. Furthermore, this study was designed as an exploratory pilot, aiming to pave the way for future research with larger and more diverse participant groups. However, we acknowledge that the relatively small sample size may limit the generalizability of our findings. The insights gained from this study provide a foundation for further research, but should be interpreted with caution when extrapolating to broader populations. Future studies should aim to involve a larger and more diverse sample to validate these initial findings and to explore variations in user experience between different demographic groups.

\subsubsection{User Experience Questionnaire}
The User Experience Questionnaire (UEQ), a standardized tool for evaluating user perceptions across various dimensions, was used to assess the participants' experiences. The UEQ was the primary statistical support for our experiment; however, we also requested direct feedback from the volunteers on their experience. UEQ is a 26-question survey that seeks to quantify the user’s experience with a product or service in six categories; attractiveness, perspicuity, efficiency, dependability, stimulation, and novelty. This survey has been used in many experiments, including those involving MHCI. The UEQ was chosen for its wide applicability, as well as the pre-existing analysis tools and background research that surrounds it. This survey allows for statistical analysis of the feasibility of a product (in this case, the general experience of operating the quadruped using gestures) as well as a comparison between the two gestures. Only the research team had access to the data, and it was used solely for the purposes of this study. No identifying information was recorded or reported in the analysis, ensuring the confidentiality of the participants. The details of the procedure is as follows: 

\subsubsection{Procedure}

\begin{itemize}
    \item The study participant was introduced to the Go1 Pro, as well as the technology they would be working with for the experiment. They were shown the informational guides to help them learn the correct movements to navigate the robot, and were given a brief introduction on its best practices when operating it. The research procedures were designed to ensure the safety and comfort of the participants throughout the experiment. Prior to the commencement of the study, participants were fully informed about the nature, purpose, and procedures of the experiment.  Any potential discomfort was discussed with participants prior to their involvement, and appropriate measures were taken to address any concerns. The study participant may also be referred to as “the user”.
    \item Following this introduction, the user was given up to four minutes to familiarize themselves with controlling the robot. Up to two researchers observed and monitored the user and software/hardware during the experiment. At this stage, the investigators would answer any questions the user may have about the technology.
    \item Following the four minutes, or when the user says they are ready, the investigators placed the Go1 Pro at the start of the course, which was a red line drawn on the floor in view of the user. The investigators told the user to complete the course using the controls at their disposal as quickly as possible. This involved walking the Go1 Pro from point-to-point, roughly following the red line. The investigators told the user that they would be timed.
    \item Following completion of the course, the investigators would offer the user the survey, a UEQ in electronic format. The users were told to answer quickly and honestly.
    \item Following completion of the first survey, the user would repeat the above process again with the alternative form of control they did not use initially. For example, if they started with body control first, they would then use hand control, and vice versa. Additionally, if User A started with body control, User B would start with hand control, to mediate potential bias created by only starting with one form of control between participants.
    \item Following completion of both methods of control and their respective survey, the user was asked which method they preferred, and why. The investigators recorded this feedback. 
\end{itemize}

To mitigate bias in the results, it was crucial to introduce all participants to the technology in a consistent manner. Although we did not hinder users from asking questions, investigators stuck to similar scripts when teaching the user how to operate the technology.

\section{Results}

With the experiment setup and procedures established, we now present the findings, focusing on user preferences and performance metrics across the different gesture types.


\begin{figure}[tb]
  \centering
  \includegraphics[height=8cm]{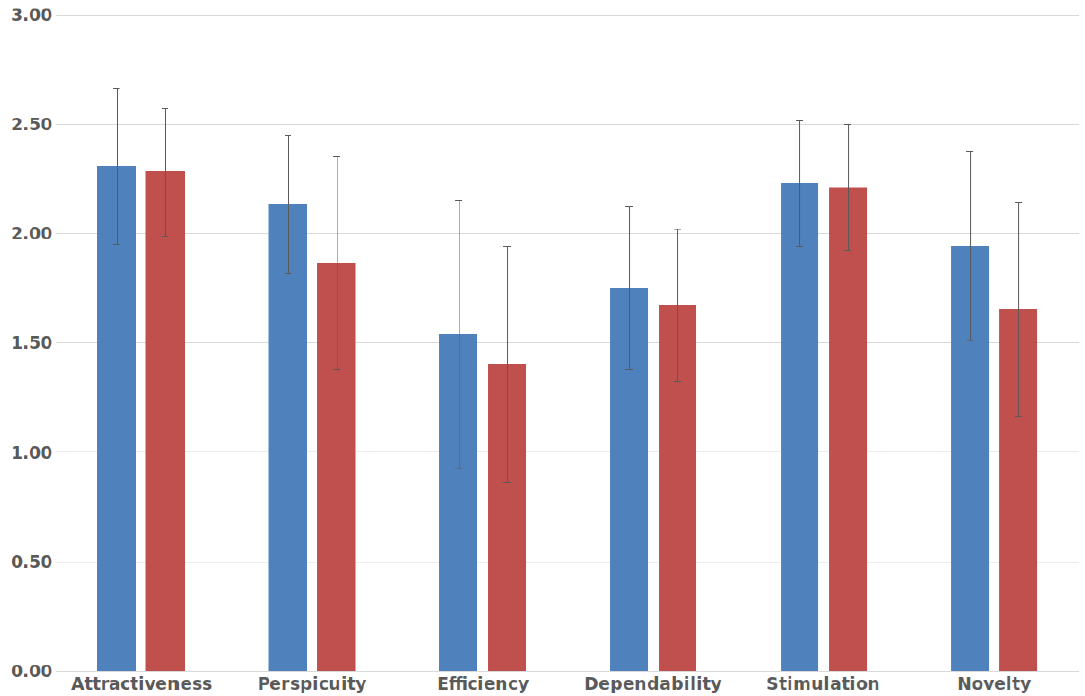}
  \caption{Comparing hand- and body-based control, where blue indicates body, and red indicates hand.
  }
  \label{fig:hand_body_comparison}
\end{figure}

\subsection{Comparison of Preference of Hand and Body}

Figure \ref{fig:hand_body_comparison} is the results on the comparison of hand and body controls, where one can see that body-based control, on average, achieved a higher score than that of hand-based control in all categories, to varying degrees, with the largest differences being in novelty, perspicuity, and efficiency. Other qualities, such as stimulation as attractiveness, are almost identical. However, the question of how significant the difference is between user satisfaction between body- and hand-based control should be asked. On the one hand, the fact that body control ended up higher for all six categories should be noted. However, running a t test between the same characteristics of each type of control with an alpha of 0.05 does not produce significant differences for each category.

Despite these differences, however, both methods of controlling the robot produced similar results in which qualities were rated higher. Attractiveness, stimulation, and perspicuity were rated higher across the board, compared to dependability and efficiency. Users found the act of controlling the Go1 Pro using their body and hands to be a relatively positive experience overall, and they found themselves able to adapt to the new controls. 

The lower scores for efficiency and dependability also correlate with what was observed during experimentation. As an aspect of the software used to control the Go1 Pro, users were required to wait about two seconds between issuing commands, lest the Go1 Pro potentially freeze. This was an aspect of the experiment users had the most trouble with, as they were required to wait long enough between commands. Additionally, this issue occurs with both methods of control, explaining why both of these characteristics are about equally as low.


\begin{figure}[tb]
  \centering
  \includegraphics[height=3.5cm]{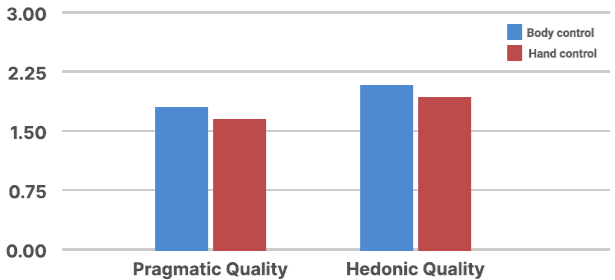}
  \caption{Comparing hand- and body-based control in terms of pragmatic and hedonic quality, where blue indicates body, and red indicates hand.
  }
  \label{fig:hedonic_quality}
\end{figure}

\subsection{Comparison of Pragmatic and Hedonic Qualities}
The various characteristics measured by the UEQ can be broken down into pragmatic qualities and hedonic qualities, as shown in Figure \ref{fig:hedonic_quality}. Simply put, pragmatic qualities define how effective the product is at accomplishing its task, and hedonic qualities define how the user feels while using the product. Efficiency, perspicuity, and dependability are pragmatic qualities, while stimulation and novelty are hedonic qualities. Attractiveness is the general user experience.

When the qualities measured by UEQ (except for attractiveness) are combined into this binary, we see that on average, users rate the hedonic qualities more positively than the pragmatic ones. While statistical significance can break down when looking at the results of individual items in the survey, one standout is the results of the slow/fast dichotomy, with this individual item being rated the lowest, on average, among all 26 items. One user, upon completion of the experiment, noted that they could have completed it a lot faster with a controller, putting words to what the data showed. 

Regarding direct user feedback following both experiments, when asked about preference, the user’s personal enjoyment, or the “fun” they may have had when operating the robot, was sometimes mentioned when selecting their preference of hand and body, as well as what was intuitive and easy.

\subsection{Comparison of Time to Task Completion}
Finally, it is worth looking at the times the participants took to complete the course.


\begin{table}[t]
    \centering
    \caption{Efficiency measurement results.}
    \label{lbl_efficiency_measurement}
    \begin{subtable}{.45\linewidth}
        \centering
        \caption{Task completion Time.}
            \begin{tabular}{|l|r|} \hline
            Category & Avg. Time \\ \hline
            Body & 3:13 \\ \hline
            Hand & 3:26 \\ \hline
            \end{tabular}
    \end{subtable}%
    \begin{subtable}{.45\linewidth}
        \caption{Task completion time per attempt.}
        \centering
            \begin{tabular}{|l|r|} \hline
            Iteration & Avg. Time \\ \hline
            First time & 3:32 \\ \hline
            Second time & 3:13 \\
            \hline
            \end{tabular}
    \end{subtable}%
\end{table}

Table \ref{lbl_efficiency_measurement} shows the results of the efficiency measurement in terms of mode and number of attempts. Two trends are immediately noticeable in these data. On average, users completed the course faster with their body than with their hand, by about 13 seconds. Additionally, there is a significant decrease in the average time between the first and second runs conducted by the users, regardless of the control type. This result further supports the validity of switching which method of control the users would start with between participants to mitigate potential bias in the final results. Additionally, fast time for body-based control supports the UEQ results, where the control type was generally more agreeable among users than that of the hand. Note that only one outlier existed when checking the IQR of each dataset, that being an upper-end outlier of the hand dataset.

Throughout the experiment, participants showed marginally higher satisfaction and completed tasks slightly faster when using body gestures to control the robot. This finding challenges the conventional emphasis on hand-based gesture control technologies for quadruped robots. It also highlights the considerable potential for advancing and refining interaction and interfacing methods, especially in contexts where collaboration and teamwork with robots are involved.





\section{Conclusion}
The results indicate a clear preference for body gestures over hand gestures, particularly in terms of user satisfaction and task efficiency. These findings have significant implications for the design of human-robot interaction systems. Traditionally, robots are often associated with control commands rather than communication, leading to the perception that "robot" typically brings to mind notions of control. However, our research in natural settings reveals that robots coexisting in human environments do not conform to this notion. 

This study aimed to explore responses to controlling robots through non-verbal and non-digital methods. Experimental results demonstrate that using body and hand gestures to interact with robots provides participants with a more engaging experience. Furthermore, bodily gestures outperform hand gestures in terms of efficiency and usability. For social robots endowed with intelligence and autonomy, it is crucial to consider interaction and gesture-based user interface/user experience (UI/UX) perspectives. 

One limitation of this study is the relatively small sample size, which may affect the generalizability of the findings. Moreover, the study's reliance on a single robot model limits the applicability of the results to other robotic platforms. Future research should investigate the scalability of these findings across different types of robots and explore more complex interaction scenarios to validate and extend these insights.


\section*{Acknowledgements}
This work is financially supported by the Korean Ministry of Land, Infrastructure and Transport (MOLIT) as the Innovative Talent Education Program for Smart City, and supported by Korea Institute for Advancement of Technology (KIAT) grant funded by the Korea Government (MOTIE) (P0020536, HRD Program for Industrial Innovation). The second and third authors were participating in this work were supported by the National Science Foundation Grant 2236524 NSF IRES Track 1: Future Mobility for Smart City. The Institute of Engineering Research at Seoul National University provided research facilities for this work.

%
%
\bibliographystyle{splncs04}
\bibliography{main}

\end{document}